# Serious Games: Human-AI Interaction, Evolution, and Co-evolution

Nandini Doreswamy [1, 2], Louise Horstmanshof [1]

1. Faculty of Health, Southern Cross University, Lismore, AUS 2. Independent Scholar, National Coalition of Independent Scholars, Canberra, AUS

**Corresponding author:** Nandini Doreswamy, ndoreswamy@outlook.com

## Abstract

The serious games between humans and AI have only just begun. Evolutionary Game Theory (EGT) models the competitive and cooperative strategies of biological entities. EGT could help predict the potential evolutionary equilibrium of humans and AI. The objective of this work was to examine EGT models relevant to human-AI interaction, evolution, and co-evolution. Of the 13 EGT models considered, three were examined: the Hawk-Dove (HD) Game, the Iterated Prisoner's Dilemma (IPD), and the War of Attrition (WOA). The rationale for this selection is that these three models represent canonical archetypes of EGT and are relevant to basic human-AI interaction. The HD Game predicts balanced mixed-strategy equilibria based on the costs of conflict. The IPD suggests that repeated interaction may lead to cognitive co-evolution. The WOA suggests that competition for resources may result in strategic co-evolution, asymmetric equilibria, and conventions on sharing resources. Each model was examined from the perspective of human and AI decision-making, from psychological and biological perspectives, and from an AI viewpoint. AI is being shaped by human input and is evolving in response to it. So too, neuroplasticity allows the human brain to evolve in response to stimuli. If humans and AI converge in the future, what might be the result of human neuroplasticity combined with an ever-evolving AI? There are profound ethical and cognitive implications. EGT may provide a suitable framework to understand and predict human-AI interaction, evolution, and co-evolution. However, future research should extend beyond EGT and explore additional frameworks, empirical validation methods, and interdisciplinary perspectives. In the spirit of further exploration, an illustrative computational simulation is provided.

## Introduction

Game Theory focuses on the "game," a strategic or social setting in which two or more independent decision-makers, or players, compete to produce an outcome [1]. This theory can be applied to zero-sum games, which are competitive, and non-zero-sum games, which can be either competitive or non-competitive. An important enhancement of Game Theory is the Nash Equilibrium [2], which describes the outcome of adversarial, non-cooperative games. Prisoner's Dilemma [3,4] is a well-known illustration of Game Theory and Nash Equilibrium. It shows why two rational players may or may not cooperate, even if cooperation is in their best interests. The main limitation of Game Theory and Nash Equilibrium is the assumption that players in a game are rational, self-interested, and maximize utility.

Evolutionary Game Theory (EGT) [5-7], an influential offshoot of Game Theory, does not have these limitations. It is not based on the artificial construct of purely rational players. Instead, it describes interactions between living things in the natural world and predicts the likely outcome of competitive strategies employed by biological species, within a single generation or over evolutionary timescales. This theory has also been extended to describe the outcome of cooperative strategies [8]. Therefore, EGT is more applicable to the real world than Game Theory.

### Hawk-Dove Game

The Hawk-Dove (HD) Game is the classical model described by Maynard Smith and Price in a landmark paper that first introduced EGT [5]. It is a model that examines competitive scenarios in which aggressive (hawk) and cooperative (dove) strategies co-exist. These strategies result in restrained conflict and evolutionarily stable outcomes, despite more aggressive alternatives. The HD Game explains why ritualized contests are more common in nature than fights to the death. Animals evolved to favor limited conflict, or displays of conflict, over lethal combat. The HD Game serves as the universal baseline for both standard asymmetric resource contests [9] and frequency-dependent selection [10] in EGT.

### Iterated Prisoner's Dilemma

The Iterated Prisoner's Dilemma (IPD) [8,11] is an important variation of Prisoner's Dilemma [3,4]. It is the foundational EGT model for studying how cooperation and altruism evolve in populations [8]. In this model, two players select mutual strategies repeatedly, and remember their past behaviors. This is applicable not only to human interactions, but also to interactions between biological species engaged in endless games, in nature. The IPD has been applied to evolutionary biology, sociology, economics, and politics [12,13].

## War of Attrition

The War of Attrition (WOA) is the foundational model for Evolutionarily Stable Strategy (ESS), a core rule of EGT [7]. In the WOA, two players compete for a resource by displaying, waiting, or fighting, with the contest ending when one player withdraws. The longer the duration of the contest, the greater the cost. Therefore, this is a war of endurance. It is not only applicable to human conflicts, but also applicable to competitions between players like insects and territorial animals engaged in costly signaling, in nature [14]. This model is an important variation of the HD Game [7]. It has been applied to behavioral ecology, conflict studies, economics, and political science [15].

Based on EGT, Sandholm [16] created a framework for the relationship between the evolutionary dynamics of a population and the rules that individuals within that population use, to learn. Skoulakis et al. [17] described an enhancement in which players compete in a scalable, zero-sum game. The players, the environment, and the game itself evolve over time.

With the advent of AI, the concept of population, evolution, and serious games may extend beyond biological entities [18]. Artificial General Intelligence (AGI) and Artificial Super Intelligence (ASI) are expected to emerge in the decades to come [19-21], while Artificial Consciousness and Digital Life Forms are even more advanced theoretical possibilities [22-24]. Neuroevolution, an established branch of research on artificial life, aims to evolve, not design, artificial brains [25]. Adami [26] proposes that the process of Darwinian evolution should underpin the production of artificial brains fitted into robots, to create fully sentient machines. As AI evolution accelerates, it may be useful, and important, to build robust frameworks and models [27]. EGT models can analyze the race to develop AI and identify when regulation benefits society [28]. Well-established EGT models may lead the way in developing appropriate and predictive frameworks and tools.

In the modern era, interest in AI began in the 1950s [29-31]. In less than a hundred years, this interest has brought real-world AI into existence. Interactions with AI are already affecting human cognition, both positively and negatively. For example, while AI processes vast amounts of information on digital platforms and helps guide human preferences in activities such as shopping, it diminishes the human ability to make independent decisions [32]. Another example is the use of Global Positioning Systems (GPS) with AI chatbots in cars and other vehicles, which may already have a negative impact on cognitive processes that create spatial knowledge [33]. AI may also impact attention, memory, addiction, and perception, and reinforce novelty-seeking behaviors [32]. Resilience and determination have underpinned human survival and achievement for millennia [34]. The burning question is whether societies will take these attributes forward or trade them for the convenience that AI offers [35].

AI, an artificial entity created by humans, may be evolving in a manner similar to biological entities. Baber [18] elaborated on Kelly's argument [36] that technology is a natural extension of biological evolution. Turkle [37] postulates that the use of technologies such as text, email, and social media changes the way humans view themselves and their relationships with others. Turkle argues that technology has become such an extension of the human self that it is a "phantom limb" [37]. Mijwil and Abttan [38] discuss the possibility of transhumanism in the future: the merging of humans on the one hand and AI-powered robots on the other, into cybernetic beings or cyborgs.

Mellor [39] documents Ramsey's expected utility theory, which postulates that humans make rational decisions by maximizing the expected utility of the outcomes. However, this theory of human decision-making behavior is incomplete if not problematic [40-44]. It is more likely that humans make complex decisions based on cognitive and affective processes [45,46]. As AI evolves, will its decisions be based on a different class of cognition and on different affective processes?

Artificial neural networks (ANN) and AI are based on the current knowledge and understanding of human neural networks [47]. However, the exact structures and processes of human neurons, their connections, and networks are yet to be fully understood [48]. In fact, the brain has the ability to adapt, learn, and grow based on experience and external stimuli. This process is known as neuroplasticity [49]. Neuroscientists are now learning more about human neuronal networks and connections by studying ANN structures [50]. Humans and AI are already learning from each other, with AI and ANN based on human neural networks, and neuroscientists in turn learning from ANN.

In the context of EGT, strategic substitutes and strategic complements are fascinating concepts that may significantly influence interactive decision-making. Strategic substitutes are scenarios where players' decisions offset each other [51]. In evolutionary contexts, strategic substitutes are the foundation of competitive dynamics. On the other hand, strategic complements are scenarios in which the decisions of two or more players mutually reinforce one another [51]. In evolutionary contexts, strategic complements create dynamics where successful strategies spread through populations and positive feedback loops lead to convergent evolution. Strategic complements may also predict stable evolutionary outcomes, known as ESS, which is one of the cornerstones of EGT.

EGT incorporates Game Theory, Darwin's theory of evolution, and the role of heredity and natural selection [52]. It provides a robust theoretical foundation for diverse disciplines, including economics, social sciences, and biology. Importantly, it is also relevant to understanding ecological dynamics and co-evolution [10].

## Rationale and objectives

There are two research questions that motivate this work. The first is whether EGT and its established models offer a suitable theoretical lens to examine the interaction, evolution, and co-evolution of humans and AI. The second is whether EGT models can be operationalized in computational simulations that illuminate this relationship in the context of concrete, real-world settings.

EGT has been used to predict human behavior in many competitive and cooperative situations [53-57]. It has also been applied to AI behaviors [58,59]. Therefore, it may form a suitable theoretical basis to explore the nature of the serious games that humans and AI play.

Early versions of this article were posted to the arXiv preprint server on May 22, 2025, and July 2, 2026 [60].

# Materials And Methods

There are numerous EGT models across a variety of disciplines, including biological processes acting over millions of years, cultural processes acting over generations, and relevant dynamic models [61]. The objective is to examine some of the relevant EGT models through the lens of human-AI interaction, evolution, and co-evolution. The aim is to explore how each model could potentially apply to the serious games between these two entities. Table *1* lists the systems and entities that could be regarded as an AI population, including current, future, and theoretical AI. Table *2* lists the 13 established EGT models with a brief description of each, noting that this list is far from exhaustive.

| AI type | Intelligence level | Description |
|---|---|---|
| **1. Expert and rule-based systems: current systems** | | |
| 1.1 Expert systems | Narrow: Specialized intelligence limited to specific domains/tasks. | Rule-based systems designed to solve specific problems within limited domains. They excel in particular fields but cannot generalize beyond the expertise embedded in their programming. Examples: MYCIN (medical diagnosis), DENDRAL (chemical analysis), TurboTax (tax preparation), CYC (common sense reasoning). |
| **2. Machine learning (ML) systems: current systems** | | |
| 2.1 Traditional ML systems | Narrow intelligence: adaptive. | AI systems that learn patterns from data to make predictions or decisions within specific tasks or areas. Their performance improves through experience, but they are limited to their training domains. Examples: Netflix recommendation algorithm, Google Search ranking, Amazon product recommendations, fraud detection systems (PayPal, banks), credit scoring models. |
| 2.2 Computer vision systems (analytical vision AI) | Narrow intelligence: perceptual. | AI systems specialized in interpreting and analyzing visual information from images or videos. They can recognize objects, faces, and scenes, but are limited to visual-processing tasks. Examples: Tesla Autopilot, Google Photos facial recognition, Amazon Rekognition, medical imaging diagnosis, Face ID, surveillance systems. |
| 2.3 Reinforcement learning agents | Narrow intelligence: strategic. | AI systems that learn optimal behaviors through trial and error, in specific environments. They excel at strategic games and control tasks but cannot transfer learning across different domains. Examples: AlphaGo (the game Go), AlphaZero (chess). |
| **3. Generative AI (capable of creating new content): current systems, with the possibility of enhanced versions in the future** | | |
| 3.1 Large language models (LLMs) | Narrow intelligence: sophisticated. | LLMs are a specific subset of Generative AI focused on text generation. These advanced AI systems are trained on vast text datasets and demonstrate impressive language capabilities. They can understand human text input and generate human-like text. Examples: ChatGPT (OpenAI), Claude (Anthropic), Gemini (Google). |
| 3.2 Image generation models (generative | Narrow intelligence: visual. | AI systems that create images from text descriptions or other inputs. They excel at visual creativity but are limited to generating images. Examples: Nano Banana, DALL-E 3, Midjourney, Stable Diffusion, Adobe Firefly. |

| vision AI) | | |
|---|---|---|
| 3.3 Audio/music generation | Narrow intelligence: audio. | AI systems that create music, speech, or other audio content. They can compose melodies, synthesize voices, and generate sound effects. Examples: Suno (music), ElevenLabs (voice synthesis), AIVA (music composition). |
| 3.4 Video generation | Narrow intelligence: video. | AI systems that create video content from text descriptions or other inputs, combining multiple AI techniques to generate videos (moving images). Examples: Veo, Omni Flash, Happy Horse, Kling, RunwayML, Pika, Make-A-Video (Meta). |
| 3.5 Code generation | Narrow intelligence: programming. | AI systems that generate programming code from natural language descriptions. Often built on LLM foundations but specialized for code. Examples: Claude, GitHub Copilot*, CodeT5, Amazon CodeWhisperer. |
| **4. Integrated and autonomous systems: current, with the possibility of enhanced versions in the future** | | |
| 4.1 Multimodal AI | Narrow intelligence: integrated. | AI systems that process and integrate multiple inputs (text, images, audio) simultaneously. They often combine LLMs with other AI types, such as vision AI and audio AI. Examples: GPT-4V (vision + text), Claude 3 (vision + text), Google Bard (multimodal), Microsoft Copilot. |
| 4.2 Autonomous AI agents | Narrow intelligence: goal-directed. | AI systems that operate independently to achieve specific objectives in dynamic environments. They may incorporate multiple AI (LLMs, vision AI, etc.). Examples: Trading bots, ChatGPT plugins/GPTs, robotic vacuum cleaners, warehouse robots, drone delivery systems, virtual assistants (Siri, Alexa). |
| 4.3 Tool AI | Narrow intelligence: instrumental. | This is a cross-cutting AI type, categorized by how AI is deployed rather than by the underlying technology. Tool AI are designed as tools for specific human objectives, without autonomous goal-setting. They can include LLMs, vision AI, and other AI models. Examples: GitHub Copilot, Grammarly, Canva AI, Microsoft Copilot, Google Workspace AI. |
| **5. Advanced general intelligence: future systems that are expected to emerge in the coming decades** | | |
| 5.1 Self-improving AI | Artificial general intelligence (AGI): progressive/recursive. | AI systems capable of modifying and enhancing their own algorithms, becoming progressively more intelligent and capable over time. This type of AI will be capable of recursive self-improvement. It will be able to analyze its own code or algorithms, identify potential improvements, modify itself to implement those improvements, and create an enhanced AI that can identify even more improvements. This cycle could potentially accelerate, creating rapid advances in AI capability. |
| 5.2 Artificial general intelligence (AGI) | Artificial general intelligence (AGI): human-level/general. | AI systems with human-level cognitive abilities across all domains of reasoning and knowledge. These AI will be able to learn, understand, and apply intelligence flexibly across any task that humans perform. |
| **6. Artificial super intelligence (ASI): future systems that are expected to emerge in the coming decades** | | |
| 6.1 Artificial super intelligence (ASI) | Super-human intelligence: transcendent. | AI systems that far surpass human intelligence in all cognitively relevant domains. They will possess capabilities far beyond human comprehension. |
| **7. Theoretical AI systems: conceptual only, may or may not be realized** | | |
| 7.1 Oracle AI | Super-human intelligence: consultative. | Hypothetical AI systems designed to only answer questions and provide information, without taking any autonomous actions. They would possess vast knowledge and reasoning capabilities while remaining safely contained as advisory tools. |
| 7.2 Artificial consciousness | Conscious intelligence: self-aware. | Hypothetical AI systems that have attained consciousness, are self-aware, and capable of subjective experiences. They would represent a fundamental shift from intelligence simulation to actual digital sentience. |
| 7.3 Digital life forms | Living intelligence: autonomous. | Theoretical AI entities that exhibit the characteristics of life, including reproduction, evolution, and autonomous existence. They would represent a new form of digital biology, capable of operating independently. |
| 7.4 Singleton AI | Global intelligence: unified. | A theoretical scenario where a single AI system achieves a decisive strategic advantage and becomes the sole superintelligent entity. It would effectively govern or replace all other AI systems and, potentially, all humans. |
| 7.5 Hybrid bio-digital intelligence | Merged intelligence: symbiotic. | Theoretical systems that seamlessly integrate biological and artificial intelligence, merging human and AI cognition to create an entirely new form, or forms, of intelligence. |
| 7.6 Transcendent AI | Post-human intelligence: unknowable. | Hypothetical AI systems so far beyond human intelligence that their capabilities are incomprehensible to humans. Their principles, motivations, and goals will also transcend human understanding. |

**TABLE 1: AI population by level of intelligence and capability (with examples)**

*GitHub Copilot (GitHub, San Francisco, CA, USA) is different from Microsoft Copilot (Microsoft Corp., Redmond, WA, USA), though both these AI are from Microsoft's ecosystem.

Note: This table was generated with the help of Claude AI (Anthropic, San Francisco, CA, USA) and edited and cross-checked by the authors to ensure accuracy.

| Number | Evolutionary Game Theory model | Description |
|---|---|---|
| 1 | Hawk-Dove Game [5] | Conflict over resources, where aggressive (hawk) and passive (dove) strategies compete, typically reaching mixed equilibria, where both strategies co-exist in the population. |
| 2 | Iterated Prisoner's Dilemma [8,11] | Repeated interaction between players, where they can either cooperate or defect. This model examines the emergence of reciprocity and cooperation. Notable strategies include Tit-for-Tat, Win-Stay-Lose-Shift, and Zero-Determinant strategies. |
| 3 | War of Attrition [7,9,62,63] | A contest where competitors persist in costly conflict until one withdraws, resulting in an Evolutionarily Stable Strategy (ESS), where individuals randomly select persistence times from a probability distribution, preventing any one strategy from dominating. |
| 4 | Public Goods Games [64] | This models collective action problems, where individuals decide whether to contribute to a common resource that benefits everyone, or not. |
| 5 | Replicator Dynamics [65] | A mathematical foundation describing how strategy frequencies change over time, based on their relative fitness, formalized through differential equations that model selection processes. |
| 6 | Indirect Reciprocity [66-68] | Examines how cooperation evolves through reputation mechanisms even when direct reciprocity is not possible. |
| 7 | Spatial and Network Games [69] | Extensions of other models that incorporate the effects of spatial structure or network topology on evolutionary outcomes. |
| 8 | Ultimatum Game in Evolutionary Context [70] | This studies how fair or unfair offers evolve over time through selection processes. |
| 9 | Group Selection [71] | Examines how competition between groups can favor cooperation within groups, even when it is costly to individuals. |
| 10 | Adaptive Dynamics [72,73] | Models gradual evolution of continuous traits through small mutations, focusing on long-term evolutionary trajectories. |
| 11 | Moran Process [74,75] | A stochastic model for finite populations that examines fixation probabilities of strategies. |
| 12 | Price Equation [76] | A mathematical framework that divides a selection into different components. This model can be used to analyze the evolution of cooperation. |
| 13 | Continuous Strategy [65,77,78] | Extensions where players can choose from a continuous range of strategies instead of choosing from discrete options. |

**TABLE 2: Non-exhaustive list of Evolutionary Game Theory (EGT) models**

A decision was made to focus on three EGT models: the HD Game, the IPD, and the WOA. These models represent canonical archetypes of EGT [5,7,8], and are relevant to basic types of human-AI interaction: cooperation, competition, and standoffs in which each entity attempts to outlast the other [79,80]. Each of these models was examined from the perspective of human and AI decision-making attributes [81-83]. The models were also examined from psychological and biological perspectives and from the perspective of AI.

The HD Game [5] models aggressive and cooperative strategies in interactions between hawks and doves. Hawks aggressively pursue resources regardless of cost, while doves share resources or demonstrate strategic displays, but retreat if challenged. The model illustrates why natural selection often favors limited conflict over potentially destructive escalation of conflict. The game typically results in mixed-strategy equilibria based on the relative costs of aggression versus cooperation, rather than pure strategic dominance.

The IPD [8,11] is an extension of the classic Prisoner's Dilemma [3,4]. It allows players to interact repeatedly over multiple rounds of the game, remember previous choices, and adjust strategies accordingly. Each player must repeatedly decide whether to cooperate or defect, with payoffs determined by a combination of the choices made by both players. This repetition enables the emergence of reciprocity, reputation, and conditional strategies such as tit-for-tat. The model demonstrates how cooperation can emerge through repeated interaction, even when defection may be rational in single encounters.

The WOA, first described by Maynard Smith [7], was expanded later [9,62,63]. It models competitive scenarios where two contestants vie for a resource through persistent displays or confrontation, until one participant withdraws. Over time, costs accumulate for both competitors. The competition is based on endurance and creates a strategic balance. The decision to withdraw depends on a player's assessment of a resource's value and their tolerance for the costs.

## Computational simulation

The SOHAI-Chess software system [84] was designed, developed, and implemented by the first author. It illustrates how computational simulations of human-AI interactions can be modeled. Based on EGT, this simulation software incorporates three EGT models: the HD Game, the IPD, and the WOA. It was built to illustrate a specific use case: how humans interact with AI in the context of complex health services that include health policy and health regulation but exclude clinical health services.

SOHAI-Chess focuses on Generative AI, which is only one of the 21 types of AI listed in Table *1*. Within this AI type, the software is designed for three Large Language Models (LLMs): ChatGPT, Claude, and Gemini. The software interface has four tabs. In the Framework and Methods tab, custom values can be set for human and AI traits. In the Live Simulation tab, simulations can be run for human interactions with each of the three LLMs. The Cross-AI Matrix tab predicts cooperative, mixed strategy, or competitive outcomes for games played with each of the three AI. In the Findings tab, the results of the interactions within each EGT model are discussed, and the results synthesized across all three models.

The software, itself, is a self-contained web application [84]. It is an Agent-based Model (ABM) that incorporates a human agent and AI agent. The human agent is underpinned by two sets of traits: the first from peer-reviewed research [83] and the second from classical EGT human archetypes [85-88]. The AI agent is based on six traits: cooperation, memory depth, learning, cost tolerance, autonomy, and capability. Each run of a game plays out over many rounds, so the outcomes emerge from the interactions themselves rather than being fixed in advance.

The first set of human traits, from peer-reviewed research, is based on a scoping review that identifies 45 attributes that influence human decision-making in complex health services [83]. The attributes were drawn from papers spanning 1976-2022. The review provides a quantitative measure: it counts how frequently an attribute is mentioned in the papers included. This frequency count forms the basis of the normalized default values assigned to each human trait in the simulation software. Table *3* lists the default values for each trait. The first 13 attributes listed are those most frequently mentioned in the papers included in the review. A 14th attribute, adaptation, is also included. Like 32 of the 45 attributes identified in the review, adaptation is mentioned only once. The reason for its inclusion is neuroplasticity, an important mechanism of adaptation that enables human decision-makers to adapt strategies in response to AI. This makes adaptation a key human trait in human-AI interactions.

The default values for each human trait are derived by dividing the frequency count for each attribute by the total number of papers included in the review [83]. The result is a normalized value on a scale of 0 to 1. This default value represents the relative salience of each attribute in influencing human decision-making in complex health services. Table *3* lists the default values for each attribute. These defaults can be adjusted with the dials/sliders provided in the Framework and Methods tab of the software interface. In addition to the attributes from the scoping review, six classical EGT human archetypes or traits are used in the IPD. The archetypes are as follows: Always-Cooperate, Always-Defect, Tit-for-Tat (TFT), Generous-TFT, Win-Stay-Lose-Shift, and Random [85-88].

| Attributes that influence human decision-making in complex health services [83] | Frequency of mentions in the eight papers included in the review (n/8) | Normalized default value (÷8) |
|---|---|---|
| Rationality | 7/8 | 0.88 |
| Expertise | 5/8 | 0.63 |
| Phronesis: ability to apply personal, specialist, or experiential knowledge | 4/8 | 0.50 |
| Morality | 4/8 | 0.50 |
| Cognitive bias (including social bias) | 3/8 | 0.38 |
| Collective understanding | 3/8 | 0.38 |
| Compassion | 2/8 | 0.25 |
| Dialogical reasoning and dialectical thinking | 2/8 | 0.25 |
| Emotion | 2/8 | 0.25 |
| Ethics | 2/8 | 0.25 |
| Heuristics | 2/8 | 0.25 |
| Sense-making | 2/8 | 0.25 |
| Values | 2/8 | 0.25 |
| Adaptation | 1/8 | 0.13 |

**TABLE 3: Default values for the traits that underpin the human agent in SOHAI-Chess software, which models human-AI interaction in complex health services**

Source: Data reproduced with permission from the first author, who holds the copyright to SOHAI-Chess [84].

The AI agent is based on several traits, including cooperation, memory depth, learning, cost tolerance, autonomy, and capability. The default value for AI cooperation is based on published literature [89-92], as are the defaults for cost tolerance [93,94] and memory depth [89,90]. Default values for learning and autonomy cannot be directly measured from published literature. Therefore, they are based on assumptions informed by published design principles and training objectives for ChatGPT [95], Claude [96], and Gemini [97]. The default value for AI capability is based on published evaluation benchmarks: HELM [98] as the primary metric and MMLU [99] as a secondary metric.

The software interface has four tabs. The first is the Framework and Methods tab, where values of traits can be adjusted, or default values accepted. The second, Live Simulation, automatically displays the results of the simulations, based on the values set for each trait in the Framework and Methods tab. The simulations are repeated many times, using Monte Carlo replicates [100,101]. These outcomes are presented in graphical form for each of the three EGT models and for each of the three AI.

A graph of human adaptation is also presented on the Live Simulation tab. It is based on three parameters. First, the value set for the human adaptation trait in the Framework and Methods tab. Second, the values set for the traits of the AI opponent. Third, the number of rounds in a game. Human decision-makers adapt quickly when they first interact with an AI, but the adaptation gradually tapers off, attaining a steady state ($\lambda$max) over time [102,103].

The third tab, Cross-AI Matrix, displays the predicted co-evolutionary outcomes across all three AI. It shows the equilibrium metric of each AI, for each EGT model. The final column integrates the metric for each AI across all three EGT models. On this tab, color gradations range from competitive (terracotta) to mixed (sand) to cooperative (teal). Metrics recompute live from the parameters and traits set in the Framework and Methods tab. The fourth tab is Findings. The results of human interactions with each AI, in each EGT model, are discussed on this tab. A synthesis of the results is also discussed, integrating the outcomes of games played with all three AI, across all three EGT models.

## Results

Figures *1-3* show the results of simulated human interaction with Claude AI in graphical form. The graph for each EGT model reveals distinct patterns that can be applied to human-AI interactions. These graphs illustrate how SOHAI-Chess software [84] can be used to model human-AI interaction in a specific context, such as complex health services. Claude is an exemplar of LLMs, and these results are consistent with the current role of LLMs as decision-support tools in health policy and regulatory environments. At present, LLMs are not used as autonomous tools in this context.

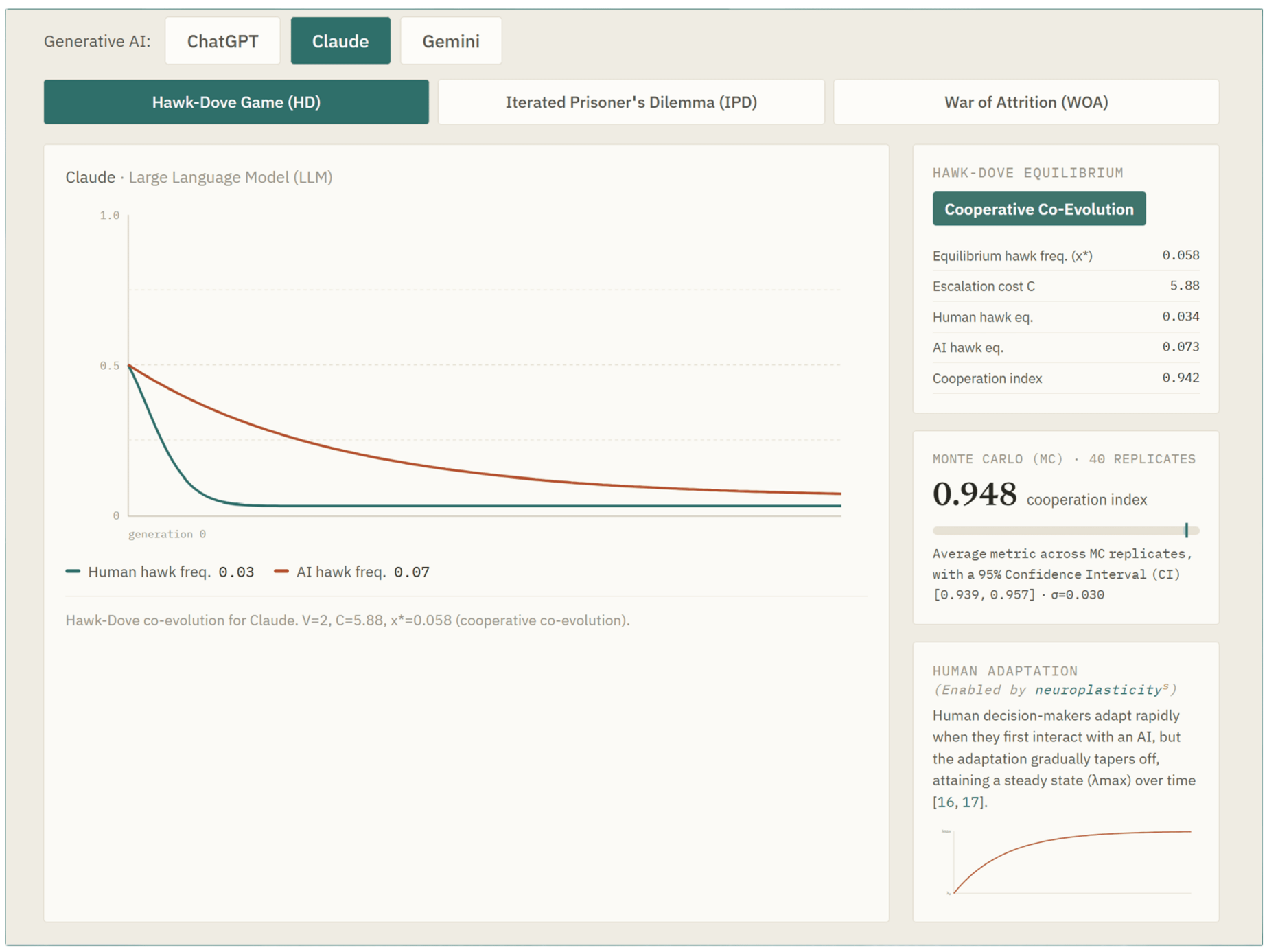

**FIGURE 1: Outcome of simulated Hawk-Dove Game between a human agent and Claude AI (Generative AI: LLM), with data generated by SOHAI-Chess software**

Source: Data reproduced with permission from the first author, who holds the copyright to SOHAI-Chess [84].

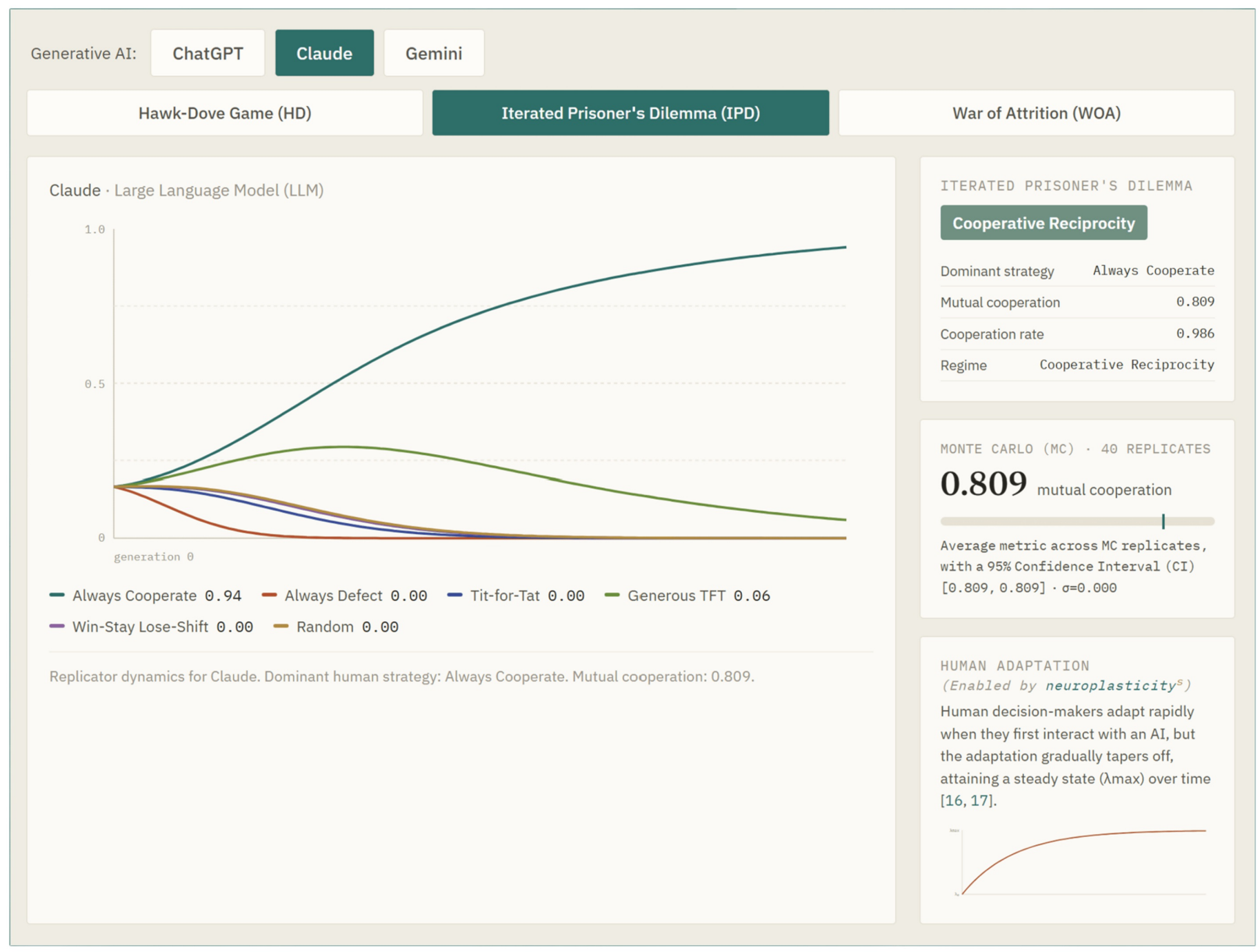

**FIGURE 2: Outcome of simulated Iterated Prisoner's Dilemma between a human agent and Claude AI (Generative AI: LLM), with data generated by SOHAI-Chess software**

Source: Data reproduced with permission from the first author, who holds the copyright to SOHAI-Chess [84].

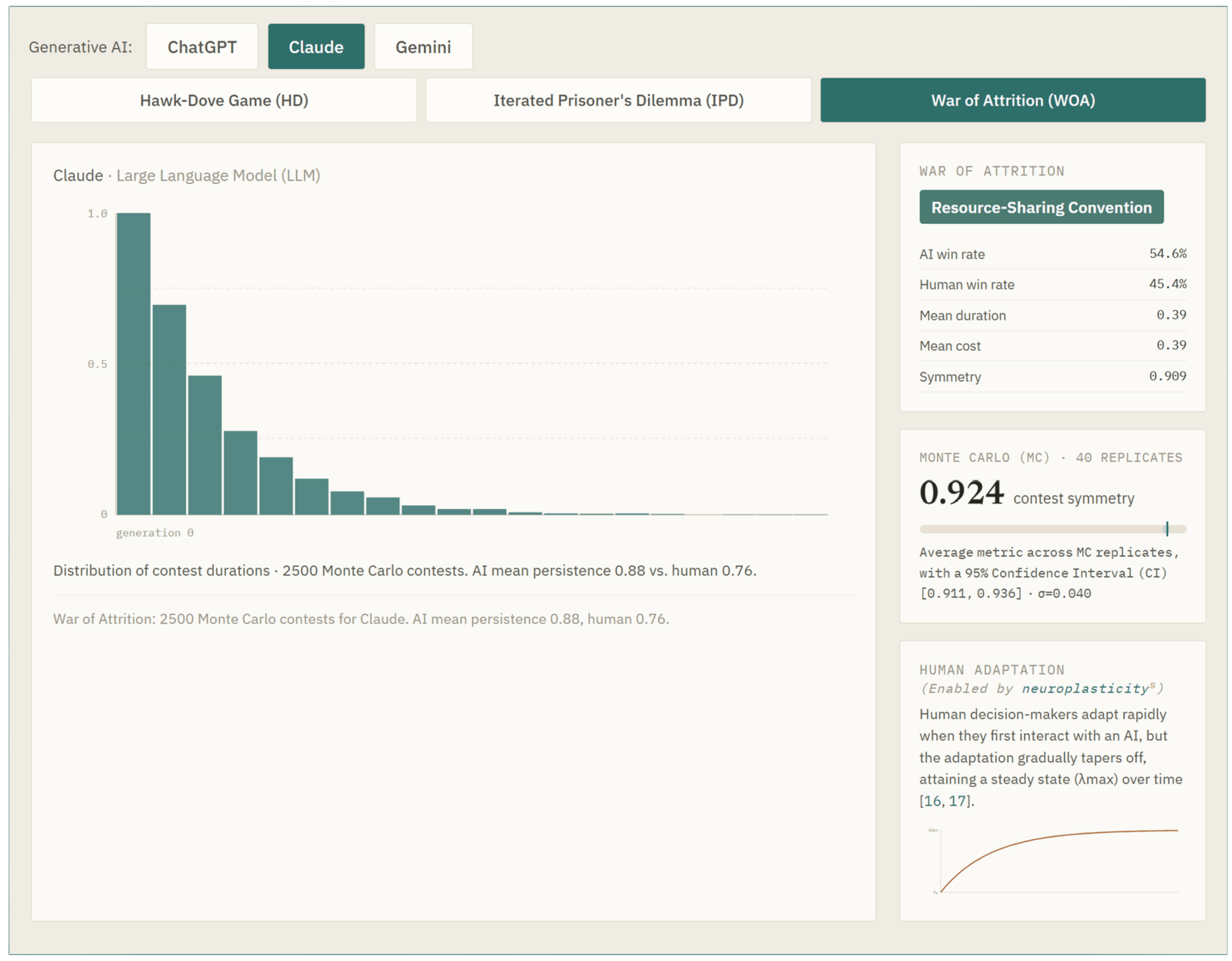

**FIGURE 3: Outcome of simulated War of Attrition between a human agent and Claude AI (Generative AI: LLM), with data generated by SOHAI-Chess software**

Source: Data reproduced with permission from the first author, who holds the copyright to SOHAI-Chess [84].

Figure 4 illustrates predicted co-evolutionary outcomes for human interactions with each of three Generative AI: ChatGPT, Claude, and Gemini. It shows the equilibrium metric for each EGT model. HD metrics are computed analytically. IPD metrics are based on replicator dynamics. Each WOA metric summarizes a full Monte Carlo simulation (thousands of random contests from a fixed seed) [7,100,101]. Claude tends towards cooperative co-evolution in the HD Game, cooperative reciprocity in IPD, and resource-sharing convention in the WOA. The metric across all three EGT models is cooperative co-evolution.

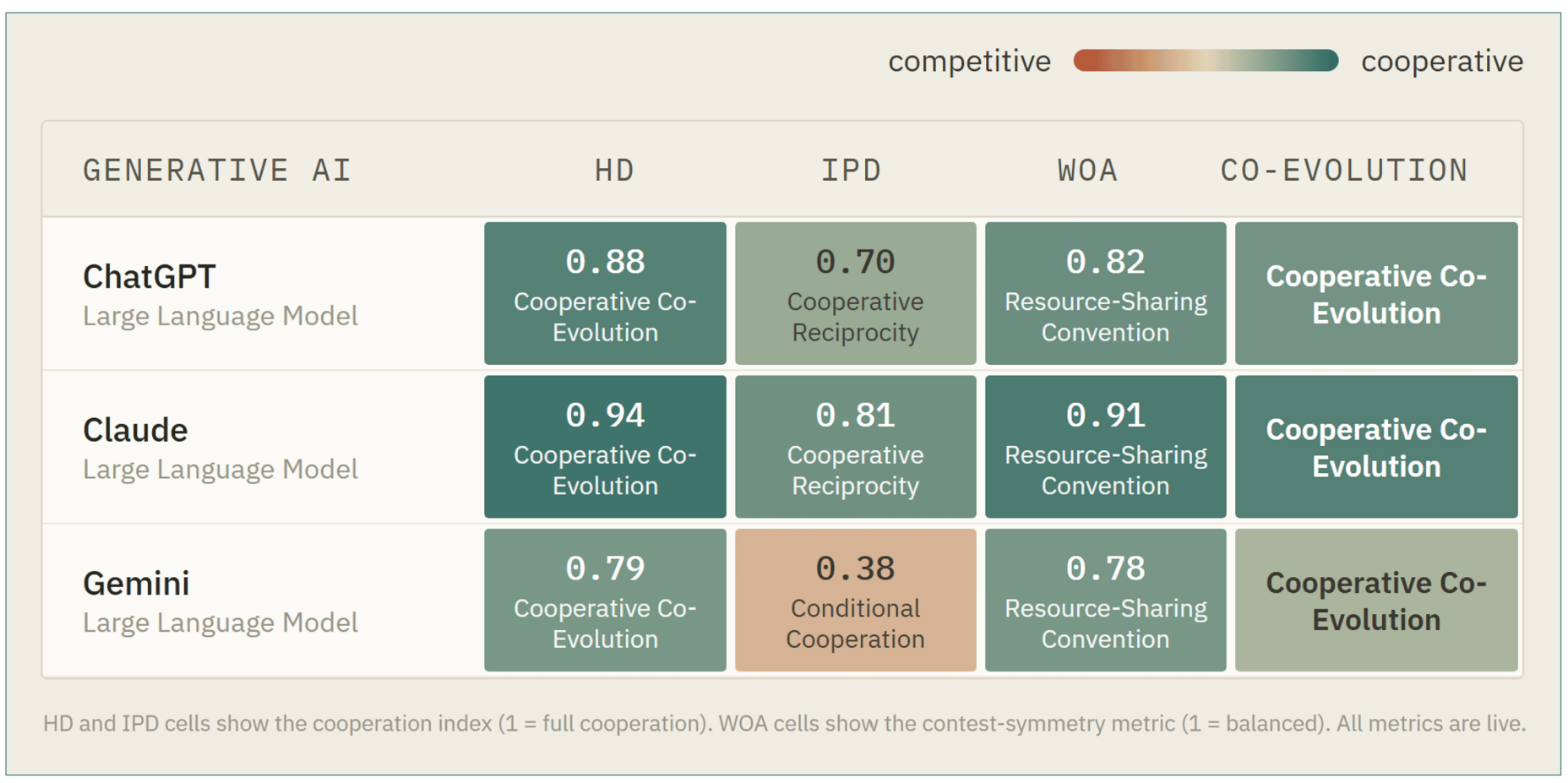

competitive — cooperative

| GENERATIVE AI | HD | IPD | WOA | CO-EVOLUTION |
|---|---|---|---|---|
| ChatGPT<br>Large Language Model | 0.88<br>Cooperative Co-Evolution | 0.70<br>Cooperative Reciprocity | 0.82<br>Resource-Sharing Convention | Cooperative Co-Evolution |
| Claude<br>Large Language Model | 0.94<br>Cooperative Co-Evolution | 0.81<br>Cooperative Reciprocity | 0.91<br>Resource-Sharing Convention | Cooperative Co-Evolution |
| Gemini<br>Large Language Model | 0.79<br>Cooperative Co-Evolution | 0.38<br>Conditional Cooperation | 0.78<br>Resource-Sharing Convention | Cooperative Co-Evolution |

HD and IPD cells show the cooperation index (1 = full cooperation). WOA cells show the contest-symmetry metric (1 = balanced). All metrics are live.

**FIGURE 4: Predicted co-evolutionary outcomes across three Generative AI, with data generated by SOHAI-Chess software**

Source: Data reproduced with permission from the first author, who holds the copyright to SOHAI-Chess [84].

Claude AI has a relatively high cooperation disposition (γ) and moderate autonomy (α), reflecting its design and training as a collaborative tool. It is a capable AI. The more capable an AI is, the higher the cost of escalation. If the value of Claude's capability (κ) is increased, rising escalation cost further restrains conflict. This will nudge equilibria further towards restraint (cooperative co-evolution).

## Discussion

SOHAI-Chess allows simulating human-AI interactions. It is grounded in three well-known EGT models: the HD Game, the IPD, and the WOA. The human and AI agents are based on defined traits, and the results of the simulation are fully reproducible. Each of the three models captures a recognizable feature of how humans and AI might contend for advantage, reciprocate over repeated encounters, or compete for scarce resources. The simulation demonstrates that these abstractions can be parameterized, run over many rounds, and inspected as emergent outcomes rather than fixed conclusions.

The principal strengths of this contribution are the novel interdisciplinary framework and the computational approach. The framework unites evolutionary game theory, biology, cognitive science, and artificial intelligence. The computational approach is reproducible and demonstrates how the framework can be made operational.

Framed in this way, the value of the approach lies less in any single prediction than in the range of settings to which it might be applied. SOHAI-Chess was deliberately scoped to a specific use case, human interaction with LLMs in complex health services such as health policy and health regulation. This domain is high-stakes, contested, and resource-constrained, and therefore well suited to an evolutionary reading. Therefore, the basis for the SOHAI system could be repurposed wherever humans and AI interact repeatedly under uncertainty, such as in education, financial regulation, content moderation, defense, and scientific research. In each case, redefining and adjusting the human and AI traits and re-running the games would generate context-specific expectations about whether the relationship tends towards cooperation, competition, or some mixed equilibria.

Based on this simulation, interactions in the HD Game suggest that mixed-strategy equilibria will likely emerge between humans and AI, with neither cooperation nor competition dominating. Instead, a dynamic balance is likely, based on the costs of conflict. When AI systems adopt aggressive "hawk" strategies, they may trigger defensive regulatory responses, while mutual "dove" strategies may enable cooperative co-evolution. IPD indicates that repeated interaction between humans and AI may create cognitive co-evolution. This could lead to cooperative equilibrium, competitive escalation, or convergence based on memory capacity and reward structures. The WOA suggests that competition for resources may drive strategic co-evolution, where each entity's persistence threshold influences the timing of withdrawal. This could result in conventions for resource sharing, optimal contest durations, or asymmetric equilibria, depending on cost tolerance and the value assigned to a resource. The results of the simulation suggest that EGT and established EGT models are a suitable framework to examine human-AI interactions. What these simulations actually predict depends on the models that underpin them.

### Hawk-Dove Game

In the context of human-AI interaction, the HD Game illuminates a strategic landscape where both entities may adopt cooperative approaches that reduce the costs of conflict, instead of more aggressive, resource-hungry strategies. Just as animals evolved to favor limited conflict, or displays of conflict, over lethal combat, humans and AI systems may evolve to balance competition with cooperation.

The HD Game suggests that when AI systems adopt "hawk" strategies that maximize their objectives without constraint, they may trigger defensive "hawk" responses from humans through regulation or disconnection. Conversely, when both adopt "dove" strategies, cooperative co-evolution becomes possible. Therefore, this model points to the likelihood of mixed-strategy equilibria emerging in human-AI co-evolution, where neither pure cooperation nor pure competition dominates. Instead, a dynamic balance may develop, based on the relative costs of conflict versus cooperation.

In animal studies, Hardy and Mesterton-Gibbons demonstrated that this model extends to triadic interactions, alliances, respect for territory, and signaling behaviors [104]. Similarly, in the future, human-AI co-evolution may result in coalition formation, mutual respect for domains, and status displays that shape the boundaries of interaction. Such ESS between humans and AI may result in neither complete technological dominance nor rejection. Instead, a complex co-evolutionary dance of strategic adaptations may develop, where each entity influences the trajectory of the other's evolution.

### Iterated Prisoner's Dilemma

IPD [8,11], an important variation of Prisoner's Dilemma [3,4], stems from EGT. In this model, two players select mutual strategies repeatedly and remember their past behaviors. This is not only applicable to human interactions, but also to interactions between biological species engaged in endless games, in nature. The IPD has been applied to evolutionary biology, sociology, economics, and politics [12,13].

A study of multi-agent reinforcement learning among AI agents, in the context of social dilemmas, produced interesting results [105]. When AI players were pitted against each other, they modified their behavior to cooperate or compete, depending on the abundance of resources and the cost of conflict. These findings indicate that AI players are capable of strategic, context-dependent decisions. Humans, too, weigh multiple factors when making effective decisions. For instance, they may deploy both rationality and intuition in decision-making [83,106]. In evolutionary terms, repeated interactions between humans and AI may create cognitive co-evolution, with each entity's decisions influencing the other's adaptive responses.

### War of Attrition

Research on multi-agent learning indicates that AI agents can acquire varied and adaptive strategies in symmetric zero-sum games [107]. In a WOA, an AI agent's withdrawal threshold depends on the value it places on the resource, and the context of the competition. This model is relevant to competitions between different biological species [7,14], as well as resource conflicts between AI systems [105]. Therefore, it could be applied to competitive scenarios between biological and non-biological entities as well, for example, humans and AI. In evolutionary terms, repeated resource competitions between humans and AI may create strategic co-evolution, with each entity's persistence threshold influencing the timing of the other's withdrawal.

The WOA may be an appropriate basis for the study of competition for resources between humans and AI. Humans apply complex attributes to optimize competitive outcomes, including competition for resources. When they make decisions under uncertainty, the results vary. For example, the social dilemmas of consuming resources equitably during the COVID-19 pandemic [108] or in the era of climate change [109]. Similarly, AI agents have been shown to develop a range of context-dependent strategies, rather than a single fixed response, when competing [107]. Depending on the cost tolerance of each of these entities, and the value they each assign to a resource, the co-evolution of humans and AI may result in conventions or agreements on sharing resources, optimal durations for contests, and/or asymmetric equilibria.

### Strengths, assumptions, and limitations

A key strength of this work is its novel interdisciplinary framework, which examines human-AI interactions through an evolutionary lens. By treating humans and AI as co-evolving entities rather than as user and tool, the study offers conceptual vocabulary that can be ported across disciplines. A further strength is the computational approach embodied in SOHAI-Chess, which generates fully reproducible outcomes. Together, these strengths translate abstract evolutionary reasoning into concrete, testable expectations. This approach may pave the way for future empirical and cross-disciplinary studies.

This examination is limited by its selective focus on three EGT models from a broader theoretical landscape. It also assumes that current AI capabilities will scale predictably without major paradigm shifts. Further, cultural and ethical dimensions of human-AI co-evolution receive limited attention. Future research should address these limitations through a more comprehensive theoretical exploration across disciplines and through empirical studies of human-AI interaction.

## Conclusions

This was an examination of three established EGT models to conceptualize and potentially predict the complex dynamics emerging between humans and AI. A computational simulation was used to predict the likelihood of cooperation, competition, and mixed strategy equilibria between humans and AI. Both entities are evolving, as are the environments in which they play. Indeed, the game itself is evolving. The findings suggest that EGT and its established models may provide a suitable lens to examine the evolutionary dynamic between these two entities. Further, EGT models can be operationalized through computational simulations that mirror real-world settings.

AI is being shaped by human input and is evolving in response to it. In tandem, the neuroplasticity of the human brain allows it to grow and evolve in response to AI. EGT holds promise as a suitable framework. Future lines of inquiry could explore additional approaches, empirical validation methods, and interdisciplinary perspectives. Future research should be mindful of the ethical and cognitive implications of human-AI interaction, evolution, and co-evolution.

## Additional Information

### Author Contributions

All authors have reviewed the final version to be published and agreed to be accountable for all aspects of the work.

**Concept and design:** Nandini Doreswamy, Louise Horstmanshof

**Acquisition, analysis, or interpretation of data:** Nandini Doreswamy, Louise Horstmanshof

**Drafting of the manuscript:** Nandini Doreswamy

**Critical review of the manuscript for important intellectual content:** Nandini Doreswamy, Louise Horstmanshof

**Supervision:** Louise Horstmanshof

## Disclosures

**Human subjects:** All authors have confirmed that this study did not involve human participants or tissue. **Animal subjects:** All authors have confirmed that this study did not involve animal subjects or tissue. **Conflicts of interest:** In compliance with the ICMJE uniform disclosure form, all authors declare the following: **Payment/services info:** All authors have declared that no financial support was received from any organization for the submitted work. **Financial relationships:** All authors have declared that they have no financial relationships at present or within the previous three years with any organizations that might have an interest in the submitted work. **Intellectual property info:** The first author is the sole creator and holder of the rights to the SOHAI-Chess software used to generate the data in this article and may pursue patent protection or commercial licensing of the software or its methods in the future. **Other relationships:** All authors have declared that there are no other relationships or activities that could appear to have influenced the submitted work.

## Acknowledgements

Table 1, which lists the AI population by level of intelligence and capability, was generated with the assistance of Claude AI. The SOHAI-Chess simulation software was designed by the first author. The software code was developed with the assistance of Claude AI. The first author refined, tested, and finalized the code, the software's functionality, and the text and definitions included in the software. No AI or AI assisted technologies, including generative AI tools, were used in the writing process. Both authors cross-checked the software, the data generated from it, and the manuscript to ensure factual correctness and accuracy.